\documentclass[letterpaper]{article} 
\usepackage[preprint]{aaai2027}
\usepackage[hyphens]{url}  
\usepackage{graphicx} 
\usepackage{natbib}  
\usepackage{caption} 
\usepackage{algorithm}
\usepackage{algorithmic}
\usepackage{multirow}

\usepackage{newfloat}
\usepackage{listings}
\DeclareCaptionStyle{ruled}{labelfont=normalfont,labelsep=colon,strut=off} 
\floatstyle{ruled}
\newfloat{listing}{tb}{lst}{}
\floatname{listing}{Listing}
\nocopyright
\usepackage{booktabs}

\makeatletter
\patchcmd{\@maketitle}
    {These authors contributed equally.}
    {Equal contribution.}{}{}
\patchcmd{\@maketitle}
    {These authors contributed equally.}
    {Equal contribution.}{}{}
\newcommand{\internshipnote}[1]{%
    \insert\aaai@thanksins{%
        \protect\footnotesize\interlinepenalty\interfootnotelinepenalty
        \splittopskip\footnotesep\splitmaxdepth\dp\strutbox
        \floatingpenalty\@MM\hsize\columnwidth\@parboxrestore
        \def\@thefnmark{}%
        \color@begingroup
        \@makefntext{%
            \rule\z@\footnotesep\ignorespaces#1\@finalstrut\strutbox}%
        \color@endgroup}}
\makeatother

\title{CAVE: Competence-Aware Visual Boundary Evidence Alignment for Video Temporal Grounding}

\author{
    Wei Jia\textsuperscript{\rm 1,\rm 2}\equalcontrib,
    Zhicong Lu\textsuperscript{\rm 2}\equalcontrib,
    Yu Chen\textsuperscript{\rm 1}\equalcontrib,
    Xiang Wang\textsuperscript{\rm 1}\corresponding,
    Shuai Li\textsuperscript{\rm 1},\\
    Wenqian Lv\textsuperscript{\rm 1},
    Jiayue Cao\textsuperscript{\rm 2},
    Huaxing Liu\textsuperscript{\rm 1}
}
\affiliations{
    \textsuperscript{\rm 1}AMAP, Alibaba Group\\
    \textsuperscript{\rm 2}University of Chinese Academy of Sciences\\
    jiawei22@mails.ucas.ac.cn, nazaritelzc@gmail.com
}

\begin{document}

\maketitle
\internshipnote{Work done during internship at AMAP, Alibaba Group.}

\begin{abstract}
Large vision-language models (LVLMs) have achieved substantial performance gains in Video Temporal Grounding (VTG) through reinforcement learning (RL).
However, existing methods primarily rely on outcome correctness rewards that evaluate only the final predicted intervals, leaving boundary-related visual evidence and its correspondence with timestamp predictions insufficiently constrained.
In this paper, we delve into timestamp prediction and its underlying boundary-level visual evidence, showing prevalent misalignment between visual evidence and predicted timestamps across widely used benchmarks.
To address this issue, we propose Competence-Aware Visual Boundary Evidence Alignment (CAVE), which augments localization optimization with boundary-specific visual evidence rewards to mitigate evidence--timestamp misalignment.
Specifically, to explicitly represent the boundary-specific visual evidence,
CAVE introduces boundary-specific evidence tokens and initializes their structured generation and distinct boundary semantics through a lightweight supervised warm-up.
During RL, the visual boundary evidence alignment reward reinforces the visual attention of special evidence tokens within the ground-truth boundaries, thereby promoting alignment between visual evidence and temporal boundaries. 
Moreover, performance-aware gating for evidence supervision is designed to adaptively retain evidence guidance for poorly localized groups while reducing it once localization becomes sufficiently accurate to avoid over-constraining fine-grained boundary refinement.
Extensive experiments on several public VTG benchmarks demonstrate the effectiveness of our method.
\end{abstract}

\section{Introduction}

Video temporal grounding (VTG) aims to precisely identify the start and end timestamps of a target event in an untrimmed video according to natural language queries. With their unified vision–language modeling capabilities, large vision-language models (LVLMs) have introduced a new generative paradigm for video temporal grounding~\cite{timesuite,chatvtg,vtimellm}.

\begin{figure}[t]
    \centering
    \includegraphics[width=0.98\columnwidth]{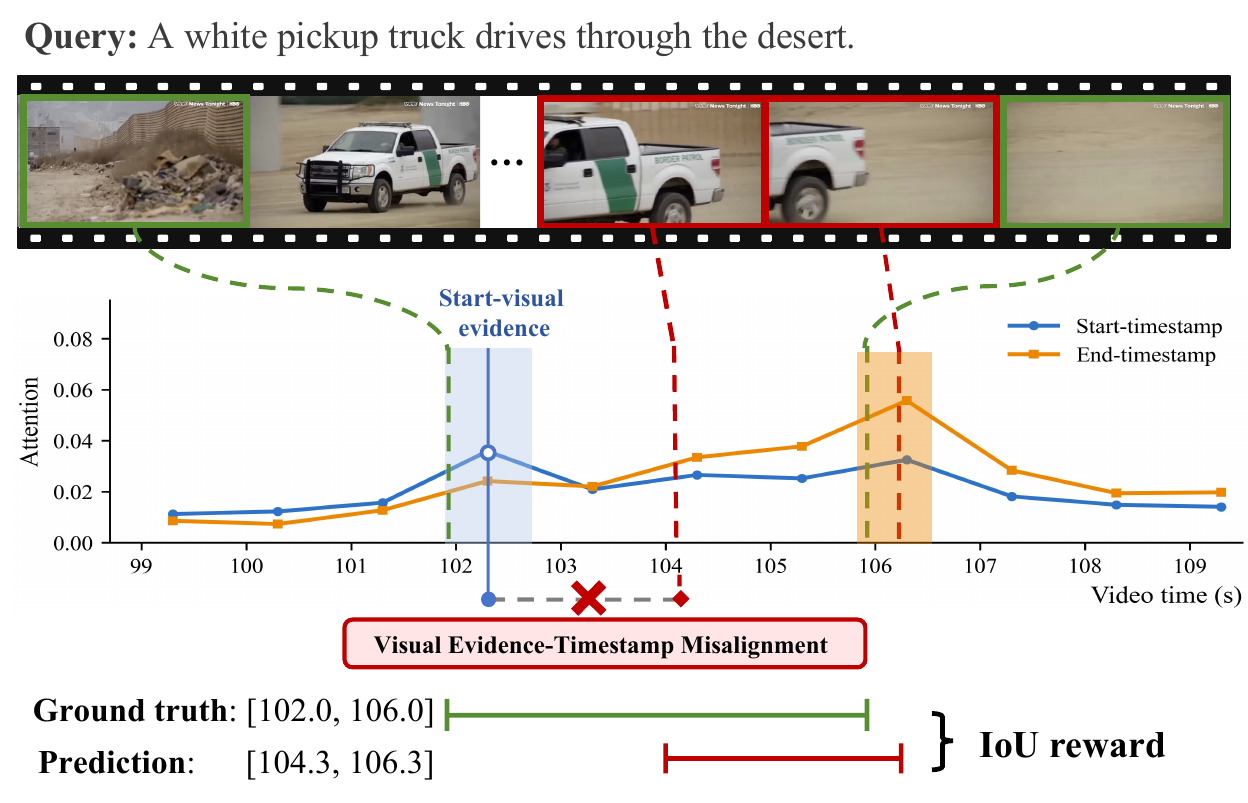}
    \caption{An illustrative example of evidence--timestamp misalignment in VTG.}
    \label{fig:intro}
\end{figure}

In recent years, reinforcement learning (RL)-based approaches have substantially improved the temporal grounding capabilities of LVLMs. 
Existing methods typically optimize verifiable rewards based on final interval accuracy~\cite{timer1,videochatr1,timelens,arrow,Datasets}, while some further construct high-quality reasoning trajectories to strengthen temporal reasoning~\cite{tartvg}.
However, such outcome-centric supervision leaves the visual evidence underlying timestamp generation largely unconstrained, leading to prevalent misalignment between predicted timestamps and boundary-related visual evidence.
As shown in Figure~\ref{fig:intro}, the generated start-timestamp token assigns its strongest attention to video frames near the ground-truth start boundary, yet the predicted start timestamp remains noticeably offset from the location indicated by this visual evidence.

To further investigate this issue, we conduct an empirical analysis of timestamp predictions and boundary-related visual attention in a representative baseline~\cite{timelens} across several widely used VTG benchmarks~\cite{timelens}.
Specifically, we compare the predicted timestamps with the model’s visual attention patterns around the ground-truth start and end boundaries.
The results show that a substantial proportion of incorrectly localized samples still exhibit concentrated attention around both ground-truth boundaries, and this phenomenon remains consistent under more stringent evaluation criteria. These findings indicate that localization outputs are not reliably coupled with the model’s internal visual responses to event boundaries, as the presence of boundary-relevant responses does not necessarily ensure that they are correctly translated into timestamp predictions. We refer to this phenomenon as visual evidence–timestamp misalignment.
Consequently, supervision based solely on temporal overlap is insufficient to constrain the correspondence between internal boundary evidence and generated timestamps, motivating the explicit modeling of their alignment.

To address this issue, we propose \textbf{C}ompetence-\textbf{A}ware \textbf{V}isual boundary \textbf{E}vidence alignment (\textbf{CAVE}), which augments timestamp optimization with boundary-specific visual rewards to alleviate evidence--timestamp misalignment and improve temporal localization.
Specifically, CAVE introduces dedicated \texttt{<Start>} and
\texttt{<End>} evidence tokens to explicitly represent visual boundary evidence. A lightweight evidence-token warm-up stage teaches the structured output format and initializes distinct boundary semantics for these tokens. 
During RL, the Visual Boundary Evidence Alignment Reward (VBEAR) reinforces the visual responses of the start and end evidence tokens within the neighborhoods of their corresponding ground-truth boundaries while suppressing irrelevant attention in non-boundary regions to promote alignment between visual evidence and temporal boundaries.
Moreover, we introduce Performance-Aware Gating for Evidence Supervision (PAGE) to further adapt evidence supervision to rollout group localization competence, retaining stronger guidance for poorly localized groups while progressively reducing it for well-localized ones to avoid over-constraining fine-grained boundary refinement.

To validate the effectiveness and generalizability of our method, 
we conduct extensive experiments on multiple representative VTG benchmarks with mainstream LVLMs. 
Overall, our contributions are summarized as follows:

\begin{itemize}
    \item We conduct an in-depth analysis of the relationship between 
    timestamp predictions and boundary-related visual responses, revealing 
    a prevalent visual evidence--timestamp misalignment in which visual 
    boundary perception is not reliably translated into timestamp outputs.

    \item We propose CAVE, which introduces dedicated boundary evidence tokens and a visual boundary evidence alignment reward to incorporate boundary-level visual supervision into timestamp optimization, thereby alleviating evidence--timestamp misalignment. A performance-aware gating mechanism further adapts this supervision to localization competence, emphasizing poorly localized rollout groups while reducing its influence on well-localized ones.

    \item Extensive experiments across multiple representative VTG benchmarks 
    demonstrate the effectiveness and generalizability of CAVE. Further analyses show that CAVE substantially improves evidence--timestamp correspondence beyond gains in localization accuracy alone.
\end{itemize}

\section{Preliminary Study}
In this section, we delve into timestamp predictions and their
boundary-related visual evidence in VTG. 
Building on prior work showing that cross-modal attention supports visual localization and that attention supervision improves grounding~\cite{attendtoevidence,guiactor}, we assess whether timestamp-token attention captures meaningful boundary cues. In a representative baseline, 
the mean attention within the ground-truth start and end neighborhoods exceeds the mean attention in 83.9\% and 82.9\% of equal-width temporal windows, respectively,
indicating boundary-sensitive and query-conditioned behavior. We therefore use timestamp-token attention to characterize boundary-related visual responses, with full diagnostics provided in the appendix.

\begin{figure}[t]
    \centering
    \includegraphics[width=0.98\columnwidth]{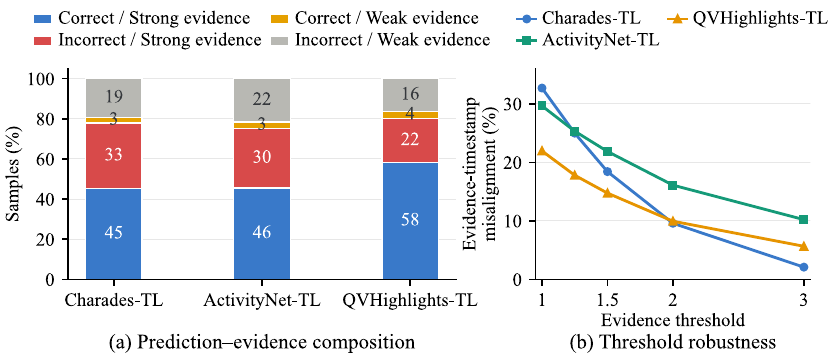}
    \caption{Diagnosing evidence--timestamp misalignment in an existing VTG
    baseline. (a) Joint distribution of localization correctness and bilateral
    boundary evidence; red denotes strict evidence--timestamp misalignment.
    (b) Strict evidence--timestamp misalignment rates under increasingly
    stringent evidence thresholds. TL denotes TimeLens.}
    \label{fig:f1}
\end{figure}

\subsection{Evidence--Timestamp Misalignment Measure}

Given a video $V=\{v_t\}_{t=1}^{T}$, a text query $q$, and a ground-truth interval $y^*=[t_s^*,t_e^*]$, a VTG model predicts an interval $\hat y=[\hat t_s,\hat t_e]$. We measure localization quality using the standard Intersection over Union (IoU) and consider a prediction correct if $\mathrm{IoU}(\hat{y},y^*)\geq\eta$.

Let $b\in\{s,e\}$ denote the start or end boundary, and let $\alpha_{b,t}^{\mathrm{num}}$ represent the final-layer attention from the corresponding generated numeric timestamp token to video frame $v_t$. For a local window $W_b$ centered at the ground-truth boundary $t_b^*$, we define $a_{b,\mathrm{win}}^{\mathrm{num}}$ as the mean attention over frames in $W_b$ and $a_{b,\mathrm{all}}^{\mathrm{num}}$ as the mean attention over the entire video. The boundary-evidence density is then defined as
\begin{equation}
D_b^{\mathrm{num}} =
\frac{a_{b,\mathrm{win}}^{\mathrm{num}}}
{a_{b,\mathrm{all}}^{\mathrm{num}}+\epsilon},
\qquad b\in\{s,e\}.
\end{equation}
A value of $D_b^{\mathrm{num}}>1$ indicates that the response around the
corresponding ground-truth boundary exceeds the video-wide average.
To require concentrated responses at both boundaries, we define strong bilateral evidence as
$\min(D_s^{\mathrm{num}},D_e^{\mathrm{num}})>\tau$,
and weak evidence otherwise.
Combining this criterion with localization correctness yields four prediction--evidence states.
We define the strict evidence--timestamp
misalignment rate at threshold $\tau$ as
\begin{equation}
\mathcal M_\tau =
\Pr\!\left(
\mathrm{IoU}(\hat y,y^*)<\eta,\,
\min(D_s^{\mathrm{num}},D_e^{\mathrm{num}})>\tau
\right).
\end{equation}
Our main analysis uses $\eta=0.5$ and $\tau=1$.
Details of boundary-window construction and attention aggregation are deferred
to the appendix.

\begin{figure*}[t]
    \centering
    \includegraphics[width=0.90\linewidth]{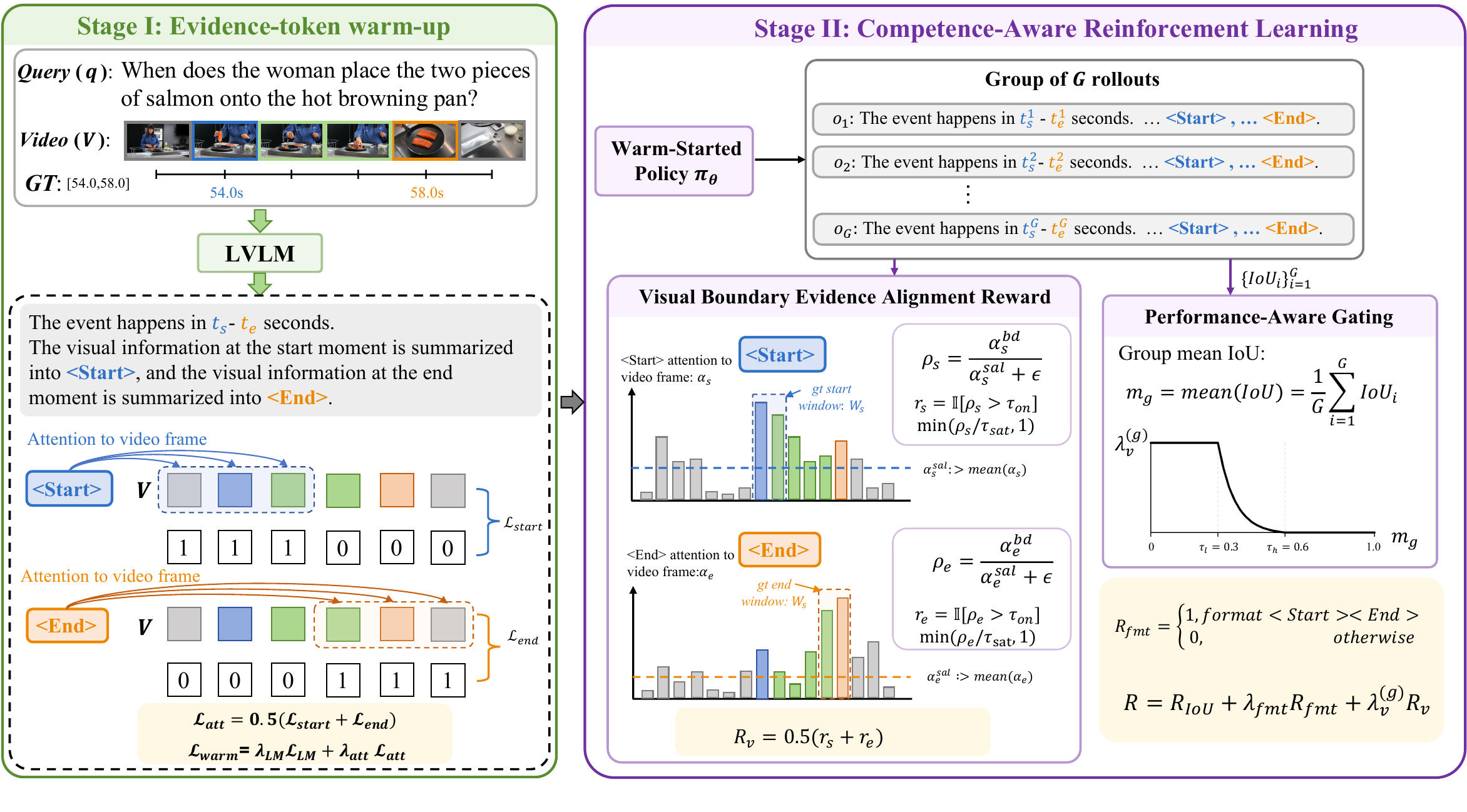}
    \caption{Overview of CAVE. A supervised warm-up initializes
    boundary-specific evidence tokens.
    During RL, VBEAR evaluates the boundary-specific visual support of each candidate interval, while PAGE adapts evidence supervision to group-level localization competence.}
    \label{fig:main}
\end{figure*}

\subsection{Exploratory Analysis}

We analyze a representative RL-based VTG baseline on three widely used
benchmarks from TimeLens~\cite{timelens}: Charades-TimeLens,
ActivityNet-TimeLens, and QVHighlights-TimeLens. 
Under its standard inference setting, we categorize predictions by localization correctness and bilateral boundary-response strength, and vary the evidence threshold $\tau$ to examine whether the observed misalignment persists under increasingly stringent criteria.

\noindent \textbf{Results and analysis.}
As shown in Figure~\ref{fig:f1} (a), approximately $22\%$--$33\%$ of valid
examples are incorrectly localized despite exhibiting strong bilateral
boundary evidence. Such cases account for more than half of all localization
failures across the three benchmarks. Meanwhile, correctly localized examples
with strong bilateral evidence do not dominate the distributions. These
results indicate that accurate timestamp prediction and boundary-related
visual responses are not reliably coupled.
Figure~\ref{fig:f1}(b) shows that the strict evidence--timestamp mismatch rate
$\mathcal{M}_{\tau}$ decreases as $\tau$ increases, but remains consistently
observable across datasets and thresholds. The mismatch is therefore not
limited to samples whose boundary responses only marginally exceed the
video-wide average. The qualitative example in Figure~\ref{fig:intro} further
shows that the start-timestamp response peaks near the ground-truth boundary
while the generated start time remains shifted, whereas the end response and
prediction are largely consistent. This suggests that a model may perceive an
individual boundary transition without reliably translating it into the
corresponding numerical timestamp.

Overall, evidence--timestamp misalignment is a prevalent phenomenon rather than
the result of a small number of outliers. The challenge lies not only in
acquiring boundary-related visual evidence, but also in converting start- and
end-specific evidence into accurate timestamp predictions. Its persistence
under stricter evidence criteria further indicates that outcome-level
optimization based solely on interval overlap is insufficient to establish a
reliable correspondence between internal boundary evidence and generated
timestamps.

\section{Method}

\subsection{Overview}

In this section, we present the proposed CAVE method. As illustrated in Figure~\ref{fig:main}, 
CAVE aims to align boundary-specific visual evidence with generated timestamps.
It introduces dedicated boundary evidence tokens and initializes their structured generation and boundary-related semantics through a lightweight supervised warm-up. During reinforcement learning,
VBEAR provides a boundary-aware visual reward that encourages each evidence token to concentrate its attention on the corresponding ground-truth boundary relative to salient off-boundary regions.
PAGE further modulates the strength of this reward according to the localization competence of each rollout group, retaining stronger guidance for difficult queries while attenuating it for well-localized ones.

Specifically, CAVE structures each model response as a numerical interval
followed by two autoregressively generated evidence tokens,
$z_s=\texttt{<Start>}$ and $z_e=\texttt{<End>}$, which correspond to the start and end boundaries, respectively. 
Let $b\in{s,e}$ denote the boundary type, and define $\alpha_{b,t}$ as the normalized attention from $z_b$ to frame $v_t$, averaged over all final-layer heads and visual tokens associated with that frame.
This ordering separates temporal-coordinate generation from boundary-evidence representation. Numerical timestamp tokens primarily encode temporal values and are not explicitly designed to summarize the visual cues underlying individual boundary decisions. By contrast, the evidence tokens are conditioned on the completed temporal decision, allowing them to summarize the boundary-specific visual information associated with the predicted start and end.

\subsection{Evidence-Token Warm-up}

To initialize the generation behavior and boundary-specific semantics of the \texttt{<Start>} and \texttt{<End>} tokens while reducing the exploration difficulty caused by sparse rollout-level rewards, we employ a lightweight supervised warm-up. The language-modeling loss $\mathcal{L}_{\mathrm{LM}}$ teaches the prescribed output format and supervises ground-truth timestamp generation.
For each boundary $b\in\{s,e\}$, let $p_{b,t}$ denote the normalized attention of the corresponding evidence token to frame $t$, and let $y_{b,t}\in\{0,1\}$ indicate whether the frame lies within the local neighborhood of the corresponding ground-truth boundary. The boundary
attention loss is defined as:
\begin{equation}
\mathcal L_b
=-\frac{1}{T}\sum_{t=1}^{T}\bigl[y_{b,t}\log p_{b,t}+(1-y_{b,t})\log(1-p_{b,t})
\bigr].
\end{equation}
We average the start- and end-boundary losses and optimize the following warm-up objective:
\begin{equation}
\mathcal L_{\mathrm{att}}
=
\frac{\mathcal L_s+\mathcal L_e}{2},
\qquad
\mathcal L_{\mathrm{warm}}
=
\lambda_{\mathrm{LM}}\mathcal L_{\mathrm{LM}}
+
\lambda_{\mathrm{att}}\mathcal L_{\mathrm{att}}.
\end{equation}
The warm-up enables reliable evidence token generation while establishing an initial distinction between start and end boundary visual semantics, providing a suitable initialization for subsequent RL.

\subsection{Competence-Aware Reinforcement Learning}

Starting from the warm-up policy, we further optimize CAVE using Group Relative Policy Optimization (GRPO)~\cite{deepseekmathh}. 
For each video--query pair,
the policy samples a group of $G$ rollouts, each containing a candidate temporal interval followed by the corresponding \texttt{<Start>} and \texttt{<End>} evidence tokens.
VBEAR evaluates the boundary-specific visual evidence associated with each rollout, while PAGE adjusts the contribution of this reward according to the localization competence of the rollout group. The gated evidence reward is then
combined with outcome-level rewards for policy optimization.

\noindent\textbf{Visual Boundary Evidence Alignment Reward.}
To distinguish valid boundary evidence from visually salient but boundary-irrelevant responses, VBEAR contrasts evidence-token attention within the corresponding ground-truth boundary neighborhood against salient off-boundary regions, thereby promoting boundary-specific evidence concentration while suppressing spurious visual saliency.

For each boundary $b\in\{s,e\}$, we first map the ground-truth timestamp $t_b^*$ to the nearest sampled frame. Starting from this boundary anchor, we construct a local neighborhood $W_b$ by expanding toward the event interior. An adjacent frame is included only when it remains sufficiently similar to both the boundary anchor and the previously selected frame, subject to a maximum neighborhood size. This similarity-guided expansion preserves local semantic continuity while reducing the likelihood of crossing the event transition.
We next construct a salient off-boundary reference set $\mathcal{S}_b$. Frames outside $W_b$ whose attention scores exceed the video-wide mean are selected as candidates.

Let $\alpha_{b,t}$ denote the attention from evidence token $z_b$ to frame
$v_t$. We define the mean attention within the boundary neighborhood and the
salient off-boundary region as:
\begin{equation}
\alpha_b^{\mathrm{bd}}
=
\frac{1}{|W_b|}
\sum_{t\in W_b}\alpha_{b,t},
\qquad
\alpha_b^{\mathrm{sal}}
=
\frac{1}{|\mathcal{S}_b|}
\sum_{t\in\mathcal{S}_b}\alpha_{b,t}.
\end{equation}
The relative boundary-evidence ratio is then defined as:
\begin{equation}
\rho_b
=
\frac{\alpha_b^{\mathrm{bd}}}
{\alpha_b^{\mathrm{sal}}+\epsilon},
\qquad b\in\{s,e\},
\end{equation}
where $\epsilon$ is a small constant for numerical stability. 
The boundary-level evidence reward is defined as:
\begin{equation}
r_b
=
\mathbf{1}\!\left[\rho_b>\tau_{\mathrm{on}}\right]
\min\!\left(
\frac{\rho_b}{\tau_{\mathrm{sat}}},
1
\right),
\qquad b\in\{s,e\},
\end{equation}
where $\tau_{\mathrm{on}}=1$ and $\tau_{\mathrm{sat}}=2$ are the activation and saturation thresholds. The overall visual reward is defined as
\begin{equation}
R_v
=
\frac{r_s+r_e}{2}.
\end{equation}

This formulation rewards a boundary only when its local evidence exceeds the salient off-boundary response. Beyond the activation threshold, the reward increases with relative evidence concentration and saturates at a bounded value. The two boundaries are evaluated independently, preventing strong evidence at one boundary from compensating for insufficient evidence at the other.

\noindent\textbf{Performance-Aware Gating for Evidence Supervision.}
To preserve useful evidence guidance without hindering fine-grained boundary refinement, PAGE estimates query-specific competence from the current on-policy rollout group and adaptively scales VBEAR, retaining stronger supervision for poorly localized groups while attenuating it for well-localized ones.

For a rollout group $g$ containing $G$ candidate intervals
$\{\hat{y}_i\}_{i=1}^{G}$, we define the group competence as
\begin{equation}
m_g
=
\frac{1}{G}
\sum_{i=1}^{G}
\mathrm{IoU}(\hat{y}_i,y^*),
\qquad
\lambda_v^{(g)}
=
f_{\exp}(m_g),
\end{equation}
where $f_{\exp}$ is a normalized inverse-exponential gating function shared by
all rollouts in the group, and $\kappa>0$ controls its decay sharpness.
The gate retains the full evidence reward when $m_g\leq\tau_l$, progressively attenuates it when $\tau_l<m_g<\tau_h$, and suppresses it when $m_g\geq\tau_h$. The exact normalized formulation is provided in the appendix. Unlike a global step-dependent schedule, PAGE adapts to both query difficulty and the evolving policy state.

\noindent\textbf{Joint Policy Optimization.}
Let $R_{\mathrm{IoU}}$ denote the overlap reward between the predicted and
ground-truth intervals, and let $R_{\mathrm{fmt}}$ indicate whether the
structured timestamps and evidence tokens follow the prescribed output
format. The final reward is
\begin{equation}
R
=
R_{\mathrm{IoU}}
+
\lambda_{\mathrm{fmt}}R_{\mathrm{fmt}}
+
\lambda_v^{(g)}R_v.
\end{equation}
PAGE modulates VBEAR before group-relative advantage estimation, after which the entire completion is optimized using the standard GRPO objective.
CAVE provides boundary-level visual supervision that encourages the policy to generate timestamp predictions associated with more reliable boundary-related visual evidence.

\section{Experiments}

\subsection{Experimental Setup}

\noindent\textbf{Datasets.}
We evaluate CAVE on three VTG benchmarks using the refined annotations provided by TimeLens~\cite{timelens}: Charades~\cite{charades} (Char-TL), ActivityNet~\cite{activitynet} (ANet-TL), and QVHighlights~\cite{qvh} (QVH-TL).

\noindent\textbf{Baselines and Evaluation Metrics.}
We compare CAVE with representative proprietary and open-source multimodal models. The proprietary baselines include GPT-4o~\cite{gpt4o}, GPT-5~\cite{gpt5}, and the Gemini series~\cite{gemini}.
The open-source baselines cover three categories: general-purpose LVLMs
(MiMo-VL-7B~\cite{mimo}, Qwen2.5-VL-7B~\cite{qwen2.5vl}, and
Qwen3-VL-8B~\cite{qwen3vl}), video-oriented models
(VideoChat-Flash-7B~\cite{videochatclash} and
VideoChat-R1-7B~\cite{videochatr1}), and VTG-specialized models
(TRACE~\cite{trace}, TimeSuite~\cite{timesuite},
Grounded-VideoLLM~\cite{groundedvideollm},
Time-R1-7B~\cite{timer1}, and TimeLens-7B/8B~\cite{timelens}).

Following prior VTG studies~\cite{timelens,timer1}, we report R1@m under IoU thresholds $m\in\{0.3,0.5,0.7\}$ and mean IoU (mIoU).

\begin{table*}[t]
\centering
\small
\setlength{\tabcolsep}{1mm}
\begin{tabular}{@{}lcccc|cccc|cccc@{}}
\toprule
\multirow{2}{*}{\textbf{Model}}
& \multicolumn{4}{c|}{\textbf{Charades-TimeLens}}
& \multicolumn{4}{c|}{\textbf{ActivityNet-TimeLens}}
& \multicolumn{4}{c}{\textbf{QVHighlights-TimeLens}} \\
\cmidrule(lr){2-5}\cmidrule(lr){6-9}\cmidrule(l){10-13}
& R1@0.3 & R1@0.5 & R1@0.7 & mIoU
& R1@0.3 & R1@0.5 & R1@0.7 & mIoU
& R1@0.3 & R1@0.5 & R1@0.7 & mIoU \\
\midrule
\multicolumn{13}{@{}l}{\textit{Proprietary Models}} \\
GPT-4o
& 60.6 & 44.5 & 23.5 & 41.8
& 55.2 & 41.4 & 25.8 & 40.4
& 69.0 & 54.8 & 38.5 & 52.1 \\
GPT-5
& 59.3 & 42.0 & 22.0 & 40.5
& 57.4 & 44.9 & 30.4 & 42.9
& 72.4 & 60.4 & 46.4 & 56.8 \\
Gemini-2.0-Flash
& 66.4 & 53.5 & 27.1 & 46.7
& 62.9 & 54.0 & 37.7 & 49.3
& 76.2 & 66.4 & 48.3 & 60.8 \\
Gemini-2.5-Flash
& 68.7 & 56.1 & 30.6 & 48.6
& 66.8 & 57.5 & 41.3 & 52.5
& 78.2 & 69.4 & 55.0 & 64.3 \\
Gemini-2.5-Pro
& 74.1 & 61.1 & 34.0 & 52.8
& 72.3 & 64.2 & 47.1 & 58.1
& 84.1 & 75.9 & 61.1 & 70.4 \\
\midrule
\multicolumn{13}{@{}l}{\textit{Open-Source Models}} \\
VideoChat-Flash-7B
& 60.2 & 37.9 & 17.8 & 39.7
& 35.5 & 21.8 & 10.5 & 24.8
& 45.2 & 30.6 & 16.7 & 32.7 \\
VideoChat-R1-7B
& 51.9 & 30.8 & 11.7 & 33.7
& 35.0 & 23.9 & 11.3 & 25.0
& 29.3 & 19.1 & 9.4 & 21.5 \\
Time-R1-7B
& 57.9 & 32.0 & 16.9 & 36.6
& 44.8 & 31.0 & 19.0 & 33.1
& 65.8 & 51.5 & 36.1 & 49.2 \\
TRACE
& 37.2 & 21.8 & 9.6 & 27.1
& 43.4 & 33.9 & 22.0 & 32.7
& 49.7 & 39.1 & 28.1 & 39.0 \\
TimeSuite
& 56.3 & 35.5 & 18.0 & 38.1
& 27.1 & 17.5 & 8.6 & 19.8
& 27.1 & 16.9 & 9.9 & 21.7 \\
Grounded-VideoLLM
& 43.3 & 28.7 & 13.5 & 30.0
& 39.2 & 29.6 & 19.5 & 30.0
& 43.7 & 33.8 & 22.5 & 33.4 \\
MiMo-VL-7B
& 57.9 & 42.6 & 20.5 & 39.6
& 49.3 & 38.7 & 22.4 & 35.5
& 57.1 & 42.6 & 28.4 & 41.5 \\
\midrule
Qwen2.5-VL-7B
& 59.7 & 37.8 & 16.6 & 39.3
& 44.1 & 31.0 & 16.1 & 31.4
& 41.5 & 27.8 & 15.2 & 31.6 \\
TimeLens-7B
& \textbf{70.5} & 55.6 & 28.4 & 48.8
& 62.8 & 51.0 & 32.6 & 46.2
& 74.1 & 62.7 & 43.1 & 56.0 \\
CAVE-7B (Ours)
& 70.4 & \textbf{56.0} & \textbf{30.2} & \textbf{50.0}
& \textbf{63.3} & \textbf{51.4} & \textbf{34.0} & \textbf{47.0}
& \textbf{77.1} & \textbf{66.8} & \textbf{48.4} & \textbf{60.2} \\
\midrule
Qwen3-VL-8B
& 69.2 & 53.4 & 27.5 & 48.3
& 62.1 & 51.2 & 34.4 & 46.8
& 74.2 & 64.6 & 49.3 & 59.4 \\
TimeLens-8B
& 76.6 & 63.0 & 35.2 & 55.2
& 68.9 & 58.4 & 40.6 & 53.2
& 80.2 & 71.6 & 55.5 & 65.5 \\
CAVE-8B (Ours)
& \textbf{77.0} & \textbf{63.8} & \textbf{35.6} & \textbf{55.4}
& 68.8 & \textbf{59.1} & \textbf{42.1} & \textbf{53.8}
& \textbf{80.3} & \textbf{71.6} & \textbf{56.8} & \textbf{66.5} \\

\bottomrule
\end{tabular}
\caption{Main results on TimeLens-Bench. Best results are highlighted in bold.}
\label{tab:timelens-main-results}
\end{table*}

\noindent\textbf{Implementations.}
We use Qwen2.5-VL-7B-Instruct as the backbone and follow the interleaved textual timestamp encoding of TimeLens. 
Videos are sampled at 2 FPS with at most 448 frames. 
CAVE performs parameter-efficient adaptation using LoRA (r=4) only during the evidence-token warm-up while keeping the vision encoder frozen. 
The evidence-token warm-up is performed for two epochs on 1,500 examples randomly sampled from the open-source TimeLens-100K~\cite{timelens}, with the language-modeling and boundary-attention losses weighted by $\lambda_{\mathrm{LM}}=0.5$ and $\lambda_{\mathrm{att}}=1.0$, respectively. 
Following prior VTG post-training methods~\cite{timer1,timelens}, we train for one epoch on the same 2,500 difficulty-filtered RL examples used by TimeLens.
For each video--query pair, we sample $G=8$ rollouts and optimize the overall reward with $\lambda_{\mathrm{fmt}}=0.1$ and $\lambda_v^{(g)}=1.0$.
The learning rates for warm-up and RL are set to $1\times10^{-4}$ and $1\times10^{-6}$, respectively.
For PAGE, we use $\tau_l=0.3$, $\tau_h=0.6$, and $\kappa=4$.
For experiments with Qwen3-VL-8B-Instruct, we follow the same training pipeline while adapting the warm-up stage to 3 epochs and setting  $\lambda_v^{(g)}=0.5$.
All other optimization settings remain unchanged.
Following TimeLens, we select checkpoints using an early-stopping strategy based on validation performance.
Training is conducted on 16 AMD MI308X GPUs with the ROCm/HIP runtime.
Additional implementation details are provided in the appendix.

\begin{table*}[htbp]
\centering
\small
\setlength{\tabcolsep}{1mm}
\begin{tabular}{@{}lcccc|cccc|cccc@{}}
\toprule
\multirow{2}{*}{\textbf{Setting}}
& \multicolumn{4}{c|}{\textbf{Charades-TimeLens}}
& \multicolumn{4}{c|}{\textbf{ActivityNet-TimeLens}}
& \multicolumn{4}{c}{\textbf{QVHighlights-TimeLens}} \\
\cmidrule(lr){2-5}\cmidrule(lr){6-9}\cmidrule(l){10-13}
& R1@0.3 & R1@0.5 & R1@0.7 & mIoU
& R1@0.3 & R1@0.5 & R1@0.7 & mIoU
& R1@0.3 & R1@0.5 & R1@0.7 & mIoU \\
\midrule
(1) Warm-up $+$ GRPO
& 69.9 & 54.4 & 28.6 & 48.5
& 60.0 & 47.2 & 29.2 & 43.3
& 73.5 & 60.7 & 39.8 & 54.3 \\
(2) (1) $+$ VBEAR
& 70.2 & 55.2 & 28.4 & 48.9
& 62.6 & 50.5 & 32.3 & 46.0
& 76.3 & 66.3 & 45.6 & 58.6 \\
(3) CAVE w/o $\mathcal L_{\mathrm{att}}$
& 69.3 & 54.1 & 29.3 & 48.7
& 60.7 & 49.5 & 31.0 & 45.0
& 75.1 & 62.4 & 42.2 & 56.1 \\
(4) (2) $+$ PAGE (CAVE)
& \textbf{70.4} & \textbf{56.0} & \textbf{30.2} & \textbf{50.0}
& \textbf{63.3} & \textbf{51.4} & \textbf{34.0} & \textbf{47.0}
& \textbf{77.1} & \textbf{66.8} & \textbf{48.4} & \textbf{60.2} \\
\bottomrule
\end{tabular}
\caption{
Ablation study of CAVE-7B on the three TimeLens-Bench datasets. Rows (1)--(2)--(4) progressively introduce VBEAR and PAGE, while row (3) removes the boundary-attention loss from the evidence-token warm-up stage.
}
\label{tab:ablation}
\end{table*}

\subsection{Main Results}

Table~\ref{tab:timelens-main-results} reports the main results on the three TimeLens benchmarks. 
Overall, CAVE achieves the strongest overall performance among open-source 7B models. 
Compared with TimeLens-7B, it obtains an average improvement of approximately $1.9$ points across the twelve evaluation metrics. The consistent gains across benchmarks demonstrate that explicitly aligning boundary-related visual evidence with timestamp prediction effectively strengthens temporal grounding.

We further evaluate CAVE using the stronger Qwen3-VL-8B backbone.
CAVE-8B outperforms TimeLens-8B across all three benchmarks on nearly every metric.
These results indicate that the proposed boundary evidence alignment remains effective when scaling to stronger multimodal backbones, suggesting that the benefit is not tied to a specific model architecture.
Notably, CAVE achieves more substantial improvements under stricter localization criteria.
While loose IoU thresholds mainly evaluate whether the predicted segment covers the queried event, higher IoU thresholds require more accurate estimation of both start and end boundaries.
The consistent improvements on these boundary-sensitive metrics indicate that CAVE does not merely improve coarse event retrieval, but effectively refines temporal boundary prediction.
This observation is consistent with the objective of VBEAR, which encourages boundary-specific visual evidence to provide more reliable guidance for timestamp generation.
Moreover, the improvements are particularly pronounced on QVHighlights-TimeLens, where CAVE achieves substantial gains across both 7B and 8B settings.
This may be because QVHighlights-TimeLens places stronger emphasis on precisely identifying the temporal extent of query-relevant moments, making boundary-specific evidence supervision more valuable.
By relatively comparing boundary-neighboring evidence with salient off-boundary responses, VBEAR provides more targeted supervision than optimizing localization overlap alone.

\subsection{Ablation Study}

To assess the contribution of each component, Table~\ref{tab:ablation} presents ablation results of CAVE-7B.
(i) Starting from the warm-up initialized GRPO baseline, introducing VBEAR consistently improves performance across all three benchmarks, with larger gains on ActivityNet-TimeLens and QVHighlights-TimeLens.
This demonstrates that boundary-specific visual evidence provides complementary supervision beyond the interval-level IoU reward, which only evaluates the final localization outcome.
(ii) Further introducing PAGE brings additional improvements across all datasets and metrics.
Compared with fixed evidence supervision, PAGE adaptively adjusts the contribution of VBEAR according to the localization competence of rollout groups.
The larger improvements on stricter localization metrics indicate that such adaptive scheduling helps preserve stronger evidence guidance for inaccurate predictions while reducing unnecessary constraints after the model becomes sufficiently competent.
(iii) Removing the boundary attention loss $\mathcal{L}_{\mathrm{att}}$ consistently degrades performance.
This verifies that the warm-up stage is not merely required for learning the output format, but also plays an important role in initializing the boundary-specific semantics of the \texttt{<Start>} and \texttt{<End>} evidence tokens.
Without such initialization, subsequent evidence-aware optimization becomes less effective.
The ablation results validate the effectiveness of introducing boundary-specific visual evidence and adaptively optimizing its supervision.

\subsection{Analyses}

\noindent\textbf{Evidence--Timestamp Misalignment Analysis.}
To evaluate whether CAVE mitigates the evidence--timestamp misalignment, we conduct the evidence--timestamp consistency analysis. 
Following the protocol of the preliminary study, we compare TimeLens-7B, Warm-up + GRPO, and CAVE-7B under identical attention extraction and thresholding settings, using the attention distributions from generated timestamp tokens to video frames.

\begin{figure}[t]
    \centering
    \includegraphics[width=0.95\columnwidth]{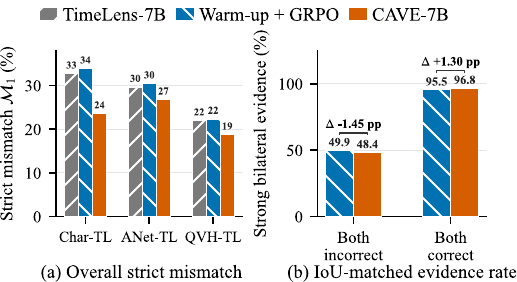}
    \caption{
    Evidence--timestamp consistency analysis using generated
    timestamp-token attention. (a) Strict mismatch rate
    $\mathcal M_1$ on three benchmarks. (b) Strong bilateral evidence
    rates for CAVE and Warm-up + GRPO on the same samples with
    $|\Delta\mathrm{IoU}|\leq0.05$, separated into pairs that are both
    incorrect or both correct. Panel (b) reports the three-dataset macro
    average, and $\Delta$ denotes CAVE minus Warm-up + GRPO.
    }
    \label{fig:evidence-gap-analysis}
\end{figure}

As shown in Figure~\ref{fig:evidence-gap-analysis} (a), CAVE achieves the lowest strict mismatch rate across all three benchmarks, indicating that it consistently reduces discrepancies between visual boundary evidence and timestamp outputs. To control for the confounding effect of improved localization accuracy, Figure~\ref{fig:evidence-gap-analysis} (b) further compares paired outputs from the same examples with nearly identical IoU and the same correctness status. Under matched localization quality, CAVE reduces the prevalence of strong bilateral evidence among incorrect predictions while increasing it among correct predictions. The former indicates fewer cases in which strong boundary responses fail to yield correct timestamps. The latter shows that accurate timestamp predictions are more often supported by clear boundary-specific visual responses. These results suggest that CAVE promotes a more reliable correspondence between visual boundary evidence and timestamp outputs, thereby contributing to more precise temporal localization.

\begin{figure}[t]
    \centering
    \includegraphics[width=0.85\columnwidth]{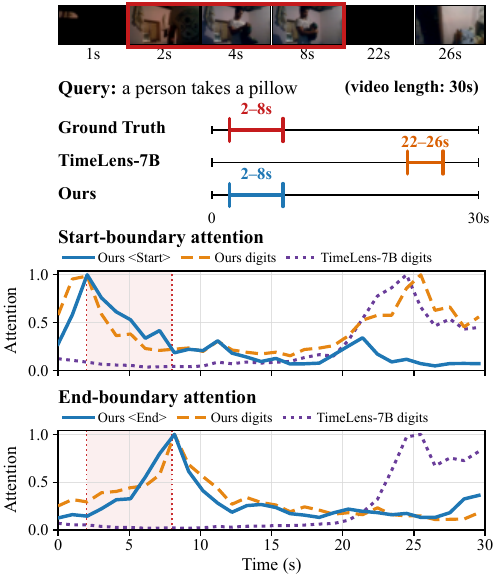}
    \caption{A case study comparing different methods.}
    \label{fig:case}
\end{figure}

\noindent\textbf{Case Study.}
Figure~\ref{fig:case} illustrates how CAVE improves the correspondence between boundary-related visual responses and timestamp predictions. 
In this case, the ground-truth interval spans $2$--$8$ seconds, whereas TimeLens incorrectly localizes a late non-target segment at $22$--$26$ seconds.
Its timestamp-token attention is likewise concentrated around this distractor rather than the queried event.
In contrast, CAVE correctly recovers the target interval. The responses of \texttt{<Start>} and \texttt{<End>}, together with the attention of the corresponding numeric timestamp tokens, peak near the respective event boundaries. This pattern is consistent with the intended effect of VBEAR, which encourages boundary-specific visual responses to concentrate around the relevant temporal transitions. This case qualitatively demonstrates that CAVE improves high-precision temporal localization while reducing evidence--timestamp misalignment.

\section{Related Work}

\noindent\textbf{Video Temporal Grounding.}
Early VTG methods typically employ pretrained visual and textual encoders to extract modality-specific features, followed by cross-modal fusion and task-specific localization modules for temporal boundary prediction. 
With the development of LVLMs, recent studies~\cite{timechat,videochatclash,timesuite,trace} increasingly formulate VTG as an autoregressive generation task and use supervised fine-tuning to directly generate temporal intervals. Several methods introduce specialized temporal or evidence tokens to improve timestamp representation~\cite{vtgllm,etbench,groundedvideollm,momentor} or event-level grounding~\cite{factor}. More recently, RL post-training has been applied to VTG~\cite{timer1,timelens,arrow} through verifiable localization rewards. Despite these advances, existing studies primarily focus on timestamp representation and final interval accuracy, while the boundary-specific visual evidence underlying the generated timestamps remains insufficiently explored.
In this paper, CAVE introduces specialized visual boundary evidence tokens to model the visual responses associated with the start and end boundaries and optimize their quality during policy learning.

\noindent\textbf{RL for Video-Language Models.}
RL has recently emerged as an effective post-training paradigm for video-language models. General video RL methods typically construct verifiable rewards from final-answer correctness or task success to improve video understanding and reasoning~\cite{videor1,videochatr1,videorft,videochatr1.5}. 
Recent studies~\cite{timer1,timelens,arrow} extend this paradigm to VTG by optimizing generated temporal
intervals with verifiable rewards based on timestamp accuracy or interval
overlap, while further exploring data quality, timestamp representation, and
thinking-free RLVR strategies.
However, these reward designs mainly evaluate the correctness of final outputs and provide limited supervision over the boundary-related visual responses underlying timestamp generation. CAVE addresses this limitation through a visual boundary evidence alignment reward that encourages boundary-specific evidence tokens to concentrate on the corresponding ground-truth boundary neighborhoods.

\section{Conclusion}

In this paper, we delve into timestamp prediction and its underlying boundary-level visual evidence in VTG, showing a
prevalent evidence--timestamp misalignment under outcome-level optimization.
To address this issue, we propose CAVE, which explicitly couples boundary-related visual evidence with timestamp generation for more reliable temporal localization. 
CAVE introduces boundary-specific evidence tokens and initializes their generation behavior and boundary semantics through a lightweight supervised warm-up.
During RL, VBEAR rewards the correspondence between boundary evidence and timestamp predictions, while PAGE modulates the strength of this supervision according to rollout group localization competence. 
Extensive experiments on representative VTG benchmarks demonstrate that CAVE improves temporal localization and effectively reduces evidence--timestamp misalignment.

\bibliography{aaai2027}

\end{document}